\documentclass[10pt,twocolumn,letterpaper]{article}

\usepackage{cvpr}              % To produce the CAMERA-READY version
\usepackage{multirow}
\usepackage{listings}   % 用于显示源代码
\usepackage{xcolor}     % 可选，支持代码高亮
\usepackage{tcolorbox}
\usepackage{enumitem}
\tcbuselibrary{breakable,skins,listings}
\usepackage{multicol}
\usepackage{tabularx}  % 支持 tabularx 环境
\usepackage{array}     % 可选，用于更灵活的列定义
\usepackage[font=small,skip=0.5mm]{caption}
\usepackage{balance}
\usepackage{flushend}

\usepackage[hypcap=false]{caption}

\newcommand{\pl}[1]{\textcolor{black}{#1}}

\definecolor{cvprblue}{rgb}{0.21,0.49,0.74}
\usepackage[pagebackref,breaklinks,colorlinks,allcolors=cvprblue]{hyperref}
\def\paperID{10208} % *** Enter the Paper ID here
\def\confName{CVPR}
\def\confYear{2026}

\title{VividFace: High-Quality and Efficient One-Step Diffusion For Video Face Enhancement}

\author{
Shulian Zhang$^{1}$\thanks{Equal contribution} \quad
Yong Guo$^{2}$\footnotemark[1] \quad
Long Peng$^{3}$\footnotemark[1] \\
Ziyang Wang$^{1}$ \quad
Ye Chen$^{1}$ \quad
Wenbo Li$^{4}$ \\
Xiao Zhang$^{5}$ \quad
Yulun Zhang$^{6}$ \quad
Jian Chen$^{2}$\thanks{Corresponding author} \\
\\
$^{1}$South China University of Technology \\
$^{2}$Max Planck Institute for Informatics \\
$^{3}$University of Science and Technology of China \\
$^{4}$The Chinese University of Hong Kong \\
$^{5}$Nanjing University of Science and Technology \\
$^{6}$Shanghai Jiao Tong University \\
}

\begin{document}

\twocolumn[{%
\renewcommand\twocolumn[1][]{#1}% 防止冲突
\maketitle
\begin{center}
    \vspace{-10pt} % 可微调与标题间距
    \includegraphics[width=1.0\linewidth]{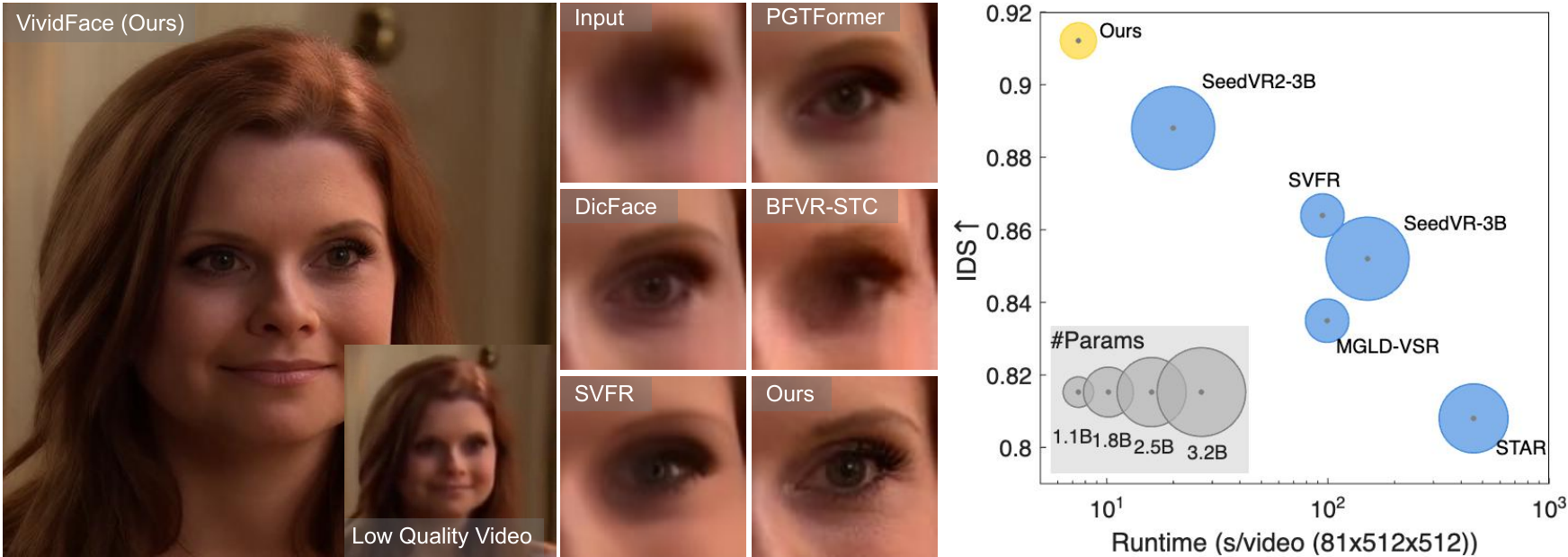}
    \captionof{figure}{
        The left side shows a visual comparison between VividFace and existing video face restoration methods,
        illustrating that VividFace produces highly realistic and visually pleasing human eyes.
        The right side compares model inference time, parameter count, and IDS performance across different methods.
        VividFace achieves best performance, fastest speed, and comparable model parameter.
    }
    \label{fig:fig1}
    \vspace{5pt} % 与下文留出适当空白
\end{center}
}]

\begin{abstract}
Video Face Enhancement (VFE) aims to restore high-quality facial regions from degraded video sequences, enabling a wide range of practical applications. Despite substantial progress in the field, current methods that primarily rely on video super-resolution and generative frameworks continue to face three fundamental challenges: (1) computational inefficiency caused by iterative multi-step denoising in diffusion models; (2) faithfully modeling intricate facial textures while preserving temporal consistency; and (3) limited model generalization due to the lack of high-quality face video training data.
To address these challenges, we propose \textbf{VividFace}, a novel and efficient one-step diffusion framework for VFE. Built upon the pretrained WANX video generation model, VividFace reformulates the traditional multi-step diffusion process as a single-step flow matching paradigm that directly maps degraded inputs to high-quality outputs with significantly reduced inference time. To enhance facial detail recovery, we introduce a Joint Latent-Pixel Face-Focused Training strategy that constructs spatiotemporally aligned facial masks to guide optimization toward critical facial regions in both latent and pixel spaces. Furthermore, we develop an MLLM-driven automated filtering pipeline that produces MLLM-Face90, a meticulously curated high-quality face video dataset, ensuring models learn from photorealistic facial textures.  Extensive experiments demonstrate that VividFace achieves superior performance in perceptual quality, identity preservation, and temporal consistency across both synthetic and real-world benchmarks. We will publicly release our code, models, and dataset to support future research.

% Built upon the pretrained WANX video generation model, our method leverages powerful spatiotemporal priors through a single-step flow matching paradigm, enabling direct mapping from degraded inputs to high-quality outputs with significantly reduced inference time. To further boost efficiency, we propose a Joint Latent-Pixel Face-Focused Training strategy that employs stochastic switching between facial region optimization and global reconstruction, providing explicit supervision in both latent and pixel spaces through a progressive two-stage training process. Additionally, we introduce an MLLM-driven data curation pipeline for automated selection of high-quality video face datasets, enhancing model generalization. Extensive experiments demonstrate that VividFace achieves state-of-the-art results in perceptual quality, identity preservation, and temporal stability. We will publicly release our code, pretrained models, and the MLLM-Face90 dataset to support future research.
\end{abstract}    
\section{Introduction}
\label{sec:intro}

Video face enhancement (VFE) aims to remove degradations and restore fine details in facial videos. It has become a fundamental technology for diverse applications such as surveillance systems, film restoration, video communication platforms, and digital content creation~\citep{wang2025bfvr,DBLP:journals/corr/abs-2107-05548,DBLP:journals/air/RotaBBS23}. The core challenge in VFE lies in accurately modeling facial textures and reconstructing realistic details while ensuring efficient processing of video sequences. Recent advances in deep learning have driven the development of numerous VFE methods~\citep{xu2024pgtformer, feng2024keep, wang2025bfvr, wang2025svfr, chen2025dicface, tan2024StableBFVR, xu2024universal, DBLP:conf/wacv/Zou0SWK25}, which have shown encouraging performance improvements.

Despite this progress, current VFE methods still face three fundamental limitations. 
First, although diffusion-based approaches have demonstrated strong generative capabilities for high-fidelity reconstruction, they suffer from significant inference efficiency bottlenecks. Their iterative multi-step sampling processes~\citep{wang2025svfr, tan2024StableBFVR, xu2024universal, DBLP:conf/wacv/Zou0SWK25} making them impractical for real-time or large-scale deployment scenarios where processing speed is critical.
Second, existing methods often struggle to recover sufficient facial details~\citep{chen2025dicface, wang2025bfvr, xu2024pgtformer}, particularly failing to reconstruct fine-grained textures in critical areas like eyes and lips. This limitation results in visually blurred or unnatural facial appearances, as illustrated in Figure~\ref{fig:fig1}. 
Third, widely used public datasets such as VoxCeleb1~\citep{nagrani2020voxceleb} and VFHQ~\citep{xie2022vfhq} exhibit inherent quality limitations, containing videos with inconsistent degradations including motion blur, poor illumination, and facial occlusions. These data quality issues fundamentally impede the effective learning of authentic facial texture structures, limiting model generalization capabilities.

% However, these methods still face several fundamental challenges. First, current methods often struggle to recover sufficient facial details, particularly failing to reconstruct fine-grained textures in key regions such as the eyes, lips, and skin, which results in blurred or unnatural facial appearances, as shown in Figure~\ref{fig:fig1}. Second, widely used public datasets like VoxCeleb1~\citep{nagrani2020voxceleb} and VFHQ~\citep{xie2022vfhq} present data quality challenges, as they contain inconsistent degradations including motion blur, poor illumination, and facial occlusions, thereby impeding the effective learning of authentic facial texture structures. Most critically, although diffusion-based approaches have demonstrated strong generative capabilities for high-fidelity reconstruction, they are hindered by significant inference efficiency bottlenecks due to their iterative multi-step sampling processes~\citep{feng2024keep,wang2025svfr,chen2025dicface}, making them computationally impractical for real-time or large-scale deployment scenarios where processing speed is crucial.

To address these limitations, we introduce \textbf{VividFace}, an efficient one-step diffusion framework for video face enhancement. Building upon the advanced text-to-video generation model WANX~\citep{DBLP:journals/corr/abs-2503-20314}, we leverage its robust spatiotemporal priors while reformulating the traditional multi-step diffusion process into a single-step paradigm using flow matching~\citep{esser2024scaling}. This reformulation enables direct and effective mapping from degraded inputs to high-quality outputs, achieving a substantial 12× speedup over SVFR~\citep{wang2025svfr} while maintaining superior visual fidelity and temporal consistency. Recognizing that facial regions contain the most perceptually critical features, we propose a Joint Latent-Pixel Face-Focused Training strategy that constructs spatiotemporally aligned facial masks to explicitly guide model optimization toward key facial areas in both latent and pixel spaces. 
Furthermore, to address the data quality challenges prevalent in existing face-centric datasets, we develop a high-quality video filtering pipeline driven by a Multimodal Large Language Model (MLLM). This pipeline automatically assesses multiple quality dimensions through carefully crafted prompts, producing our curated MLLM-Face90 dataset that ensures photorealistic facial texture learning. Extensive experiments on both synthetic and real-world datasets demonstrate the superior performance of our approach compared to existing methods.

% Specifically, we build upon the advanced pretrained WANX~\citep{DBLP:journals/corr/abs-2503-20314} video generation model to provide strong spatiotemporal priors. By reformulating the traditional multi-step diffusion process into a single-step paradigm using flow matching~\citep{esser2024scaling}, our method enables direct mapping from degraded inputs to high-quality outputs, greatly improving inference speed and efficiency. To enhance the restoration of facial details, we propose a Joint Latent-Pixel Face-Focused Training strategy that constructs facial masks aligned with the latent geometry of the VAE encoder, guiding the model to focus on key facial regions during optimization. Furthermore, we develop an MLLM-driven high-quality video filtering pipeline to automatically curate reliable face-centric training data, which helps overcome the data quality issues present in existing training datasets. Extensive experiments on both synthetic and real-world datasets demonstrate the superior performance of our approach compared to existing methods.

We summarize our contributions as follows:
\begin{itemize}
    \item \pl{We introduce the first one-step diffusion framework tailored for video face enhancement, achieving a remarkable 12× speedup over SVFR while consistently outperforming existing methods across diverse evaluation metrics on both synthetic and real-world datasets.}
    \item \pl{We propose a novel Joint Latent-Pixel Face-Focused Training strategy, which provides explicit facial guidance in both latent and pixel spaces, enabling more targeted optimization of facial regions through a progressive two-stage training process.}
    \item \pl{We develop an automated MLLM-driven high-quality video filtering pipeline and present MLLM-Face90, a meticulously curated dataset containing 1,957 high-quality face video clips, empowering the model to learn more authentic facial textures and details.}
\end{itemize}
\section{Related Work}
\label{sec:relatedwork}

\pl{\textbf{Video Face Enhancement.}}
\pl{Face video enhancement aims to recover high-quality facial video from degraded video, and finds application in surveillance, entertainment, and video communication scenarios. Directly applying general image and video enhancement methods~\citep{chan2022basicvsr++,DBLP:journals/corr/abs-2201-12288,chan2022realbasicvsr,yang2023mgldvsr,xie2025star,wang2025seedvr,li2023ntire,ren2024ninth,wang2025ntire,peng2020cumulative,wang2023decoupling} often leads to suboptimal results. Recent research has focused on dedicated face video enhancement approaches to address unique challenges such as inter-frame flickering, identity drift, and texture inconsistencies. PGTFormer~\citep{xu2024pgtformer} is the first method tailored for video face enhancement, enabling end-to-end enhancement without pre-alignment. KEEP~\citep{feng2024keep} improves temporal consistency by recursively leveraging previously restored frames to guide current frame enhancement. SVFR~\citep{wang2025svfr} utilizes generation and motion priors from Stable Video Diffusion for more robust enhancement. DicFace~\citep{chen2025dicface} introduces the Dirichlet distribution for continuous codebook combination, offering greater flexibility in representation. Furthermore, current methods are still limited by the low quality of available data. Although existing datasets such as VoxCeleb1~\citep{nagrani2020voxceleb}, CelebV-HQ~\citep{zhu2022celebvhq}, VFHQ~\citep{xie2022vfhq}, and FOS~\citep{chen2024fos} provide a large amount of facial video data, these datasets often contain degradations such as motion blur, poor lighting, and occlusions, resulting in suboptimal training data quality. These low-quality data samples pose significant challenges for face enhancement methods, making it difficult to efficiently generate realistic facial textures.}

\noindent\textbf{One-step Diffusion.}
Diffusion models~\citep{song2020score,ho2020ddpm,song2020ddim,rombach2022latentdiffusion} have achieved impressive visual results in various tasks, thanks to their ability to generate high-fidelity and realistic frames. However, their multi-step inference process leads to high computational cost and slow generation, especially for video data where efficiency is crucial. Recently, one-step diffusion methods have been proposed to accelerate generation, and have shown promising results in both image~\citep{li2024distillation,wu2024osediff,wang2024sinsr,dong2025tsdsr} and video super-resolution~\citep{chen2025dove,wang2025seedvr2,sun2025dloral,liu2025ultravsr}. Despite these advances, the application of one-step diffusion remains largely unexplored for blind face video enhancement, leaving a gap in this important area.
% \section{Methodology}
% \section{Efficient One-Step Diffusion with Face-Focused Optimization}
\section{Efficient Video Face Enhancement}
In the following, we focus on addressing the key challenges in video face enhancement: computational inefficiency of multi-step diffusion models, insufficient facial detail reconstruction, and limited training data quality. First, we reformulate multi-step diffusion into an efficient one-step flow to accelerate inference (Section~\ref{sec:overview}). Next, we introduce a Joint Latent–Pixel Face-Focused Training strategy that explicitly concentrates learning on key facial regions (Section~\ref{ssec:latent_prior}). Finally, we design an MLLM-driven high-quality video filtering pipeline to automatically curate face-centric training data with superior quality (Section~\ref{sec:dataset}).

\begin{figure*}[t]
    \centering
    \includegraphics[width=1\linewidth]{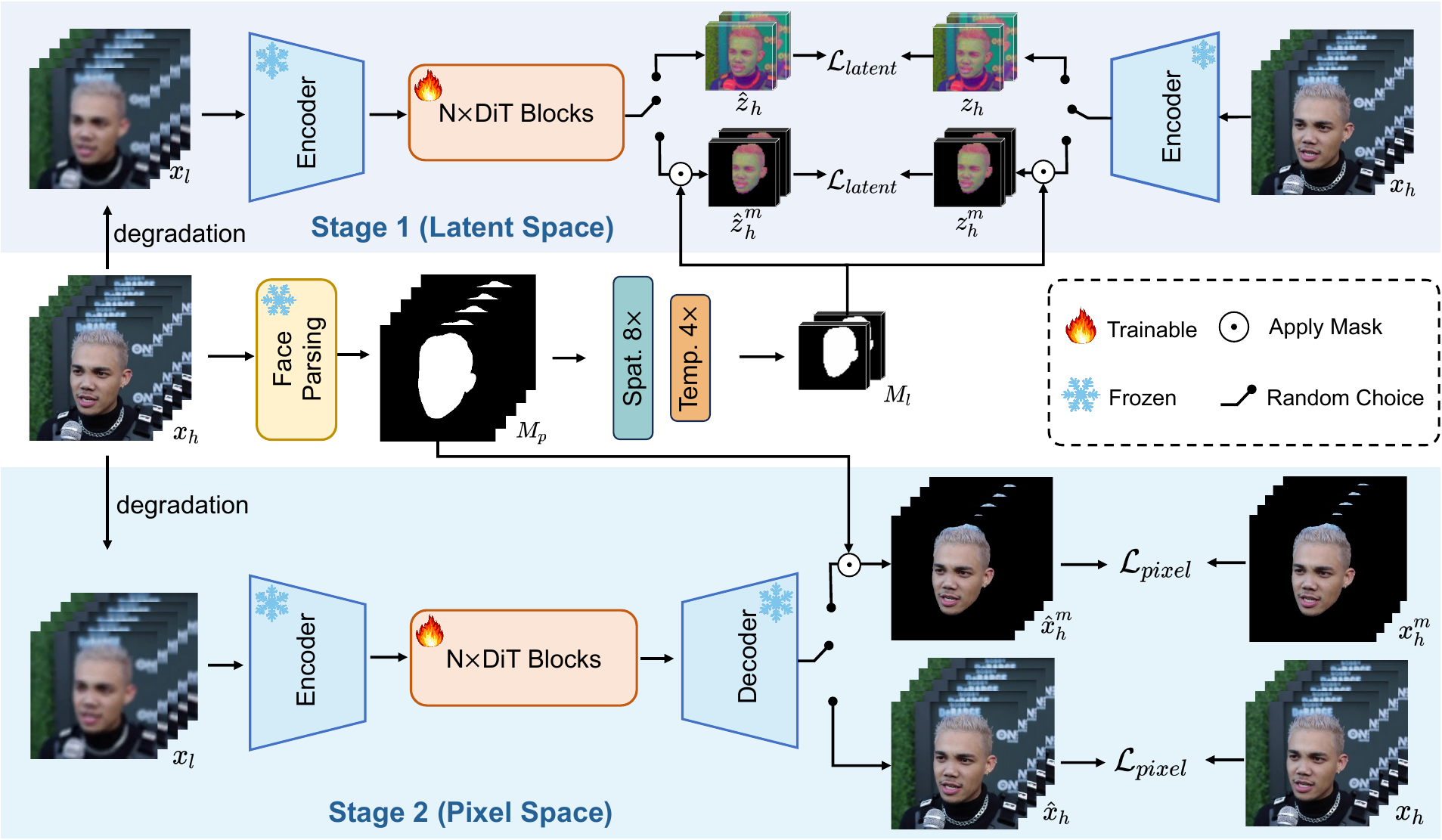} 
    \caption{\pl{Overview of our proposed VividFace training framework. VividFace is a one-step diffusion method built upon the powerful WANX video model. It adopts a two-stage design that integrates latent and pixel-space optimization, leveraging spatiotemporal priors and stochastic training to simultaneously enhance facial details and overall video quality.}}
    \label{fig:framework}
    % \vspace{-3mm}
\end{figure*}

% \subsection{One Step Flow Matching for video diffusion models}
\subsection{One Step Flow Matching}
\label{sec:overview}
Multi-step diffusion models suffer from significant computational overhead due to iterative sampling, hindering their real-time deployment. To address this bottleneck, we reformulate the diffusion process into a one-step flow matching framework built upon WANX~\citep{DBLP:journals/corr/abs-2503-20314}, a powerful pretrained text-to-video generation model. WANX employs an encoder-DiT-decoder architecture and is pretrained on large-scale real-world video datasets via multi-step flow matching, thereby providing robust spatiotemporal generative priors. Leveraging these priors, our method effectively adapts WANX to handle diverse degradation patterns in face video enhancement. Critically, since degraded face videos already retain sufficient structural information, we reformulate the multi-step process into a direct single-step transformation that maps low-quality inputs to high-quality outputs. This design achieves a remarkable 12$\times$ speedup over SVFR while maintaining superior visual quality (see Table~\ref{table:running_time}).

% We propose VividFace, a one-step face video enhancement network, leveraging the powerful pretrained text-to-video generation model WANX. WANX adopts an encoder–DiT–decoder architecture to model video generation in the latent space via flow matching, and is pretrained on large-scale real-world video datasets, enabling it to acquire robust generative priors for video enhancement. By capitalizing on generative priors, our model is able to effectively handle diverse and complex degradation conditions. Furthermore, to reduce inference time, we reformulate the multi-step flow matching into a single-step paradigm, enabling direct transformation from degraded video inputs to high-quality video outputs, achieving fast speed.

The overall architecture is illustrated in Fig.~\ref{fig:framework}. Following Rectified Flows~\citep{esser2024scaling}, given a low-quality face video $x_l$, we first encode it into a latent representation $z_l$ using the VAE encoder $\mathcal{E}$. We designate $z_l$ as the flow starting point and define a linear flow trajectory between degraded input $z_l$ and high-quality target $z_h$, as follows:
\begin{equation}
z_t = (1-t)z_l + tz_h, \quad t \in [0,1],
\end{equation}
where $t$ represents the time step. The target velocity field is then given by:
\begin{equation}
v_t = \frac{dz_t}{dt} = z_h - z_l.
\end{equation}
The DiT model $v_{\theta}$ is trained to predict this velocity field $v_t$ through a single denoising step, generating high-quality latent $\hat{z}_h$, as follows:
\begin{equation}
\hat{z}_h = z_l + v_{\theta}(z_l, t^*, c_{txt}),
\end{equation}
where $c_{txt}$ represents the text embedding and $t^*$ is the fixed timestep. 
To facilitate efficient training of the DiT, we precompute the latent representations of both low- and high-quality videos using the VAE encoder and adopt empty text prompts to eliminate caption-related computational overhead.
Empirically, we set $t^* = 400$ in the original discrete timestep scale (corresponding to $t \approx 0.4$ in the continuous scale), which balances structural preservation and detail enhancement based on the observation that the low-resolution input already contains sufficient structural information.  
Finally, the enhanced latent representation $\hat{z}_h$ is decoded by the VAE decoder $\mathcal{D}$ to produce the restored video $\hat{x}_h$.

\subsection{Joint Latent-Pixel Face-Focused Training}
\label{ssec:latent_prior}
We observe that backgrounds in facial datasets are typically blurred with limited learnable information, while facial regions contain the most visually critical features for enhancement tasks. Therefore, we propose a Joint Latent-Pixel Face-Focused Training strategy that explicitly directs the model's optimization toward key facial areas in both latent and pixel spaces.

To enable effective face-focused supervision in latent space, we construct facial masks that are spatially downsampled and temporally aggregated to match the resolution and frame rate of the VAE encoder’s latent representation.
Specifically, given an input video $x_l \in \mathbb{R}^{(1+T)\times H\times W\times 3}$, we obtain its latent encoding $z_l=\mathcal{E}(x_l)\in\mathbb{R}^{C\times T'\times H'\times W'}$, where $C=16$, $H'=H/8$, $W'=W/8$, and $T'=1+T/4$. Our goal is to produce facial masks at the latent resolution that align spatially and temporally with $z_l$. 
Instead of encoding masks directly through the VAE, which would be computationally expensive, we adopt an efficient geometric alignment strategy.
As illustrated in Fig.~\ref{fig:framework} (middle), per-frame binary facial masks $M_p\in\{0,1\}^{(1+T)\times H\times W}$ are extracted from the high-quality ground truth video $x_h$ using a face parsing model~\citep{yu2018bisenet}. We then spatially downsample these masks by a factor of 8 using nearest-neighbor interpolation to obtain: 
% $\widetilde{M}=\mathcal{D}_s(M_p)\in[0,1]^{(1+T)\times H'\times W'}$.
\begin{equation}
    \widetilde{M} = \mathcal{D}_s(M_p) \in [0,1]^{(1+T) \times H' \times W'}.
\end{equation}
Next, temporal alignment is performed to match the encoder’s temporal compression.
For latent temporal indices $i=0,\ldots,T'-1$, we set $\widehat{M}^{(0)}=\widetilde{M}^{(0)}$ for the first frame, while subsequent frames aggregate every four consecutive frames through element-wise maximum operation:
% $\widehat{M}^{(i)} = \max_{j=1}^{4} \widetilde{M}^{\,(4(i-1)+j)}, \quad i=1,\ldots,T'-1$.
\begin{equation}
    \widehat{M}^{(i)} = \max_{j=1}^{4} \widetilde{M}^{(4(i-1) + j)}, \quad i=1, \ldots, T' - 1.
\end{equation}
This process yields a spatiotemporally aligned facial mask $\widehat{M}\in[0,1]^{T'\times H'\times W'}$, which is then replicated along the channel dimension to produce $M_l\in[0,1]^{C\times T'\times H'\times W'}$ that matches the shape of the latent representation $z_l$.

With the facial masks properly constructed, we incorporate them into our training framework using a stochastic strategy that randomly switches between face-focused and global reconstruction objectives. This balanced stochastic approach proves effective, outperforming both pure face-focused and pure global training alternatives (see ablation in Table~\ref{table:mask_probability}).
Formally, for a given representation space with ground-truth $y$, prediction $\hat{y}$, and mask $M$, the loss function is defined as:
\begin{equation} \mathcal{L}(y, \hat{y}, M) = b \cdot \|M \odot (\hat{y} - y)\|_2^2 + (1-b) \cdot \|\hat{y} - y\|_2^2, \label{eq:stochastic_loss} \end{equation} where  $\odot$ denotes element-wise multiplication, and $b\sim\text{Bernoulli}(p)$ is a binary random variable that stochastically selects between face-focused supervision ($b=1$) and global reconstruction ($b=0$) with probability $p$.

To accelerate convergence and enhance optimization stability, we design a progressive two-stage training strategy that leverages the complementary strengths of latent and pixel spaces. In the first stage, the model learns to fit the one-step flow trajectory by optimizing the latent space loss:
\begin{equation}
\mathcal{L}_{\text{latent}} = \mathcal{L}(z_h, \hat{z}_h, M_l),
\end{equation} 
where $z_h$ and $\hat{z}_h$ represent the latent representations of the target and predicted, respectively, and $M_l$ is the facial mask in latent space. By operating directly in the compressed latent manifold, the model efficiently captures essential facial geometry and motion patterns before proceeding to pixel-level refinement.
In the second stage, fine-tuning is performed in pixel space using a combination of MSE and perceptual losses:
\begin{equation}
\mathcal{L}_{\text{pixel}} = \mathcal{L}(x_h, \hat{x}_h, M_p) + \lambda \mathcal{L}_{\text{DISTS}}(x_h, \hat{x}_h),
\end{equation}
where $x_h$ and $\hat{x}_h$ are the ground-truth and predicted RGB images, $M_p$ is the facial mask in pixel space, and $\mathcal{L}_{\text{DISTS}}$~\citep{DBLP:journals/corr/abs-2004-07728} is a perceptual loss that enhances fine-grained detail generation. 
This hierarchical optimization leverages latent-space learning efficiency to establish robust structural foundations, followed by targeted pixel-space refinement that enhances perceptual fidelity, resulting in superior restoration quality (see effectiveness in Section~\ref{ssec:sota}).

\begin{figure*}[ht]
    \centering
    % \vspace{2mm}
    \includegraphics[width=1\linewidth]{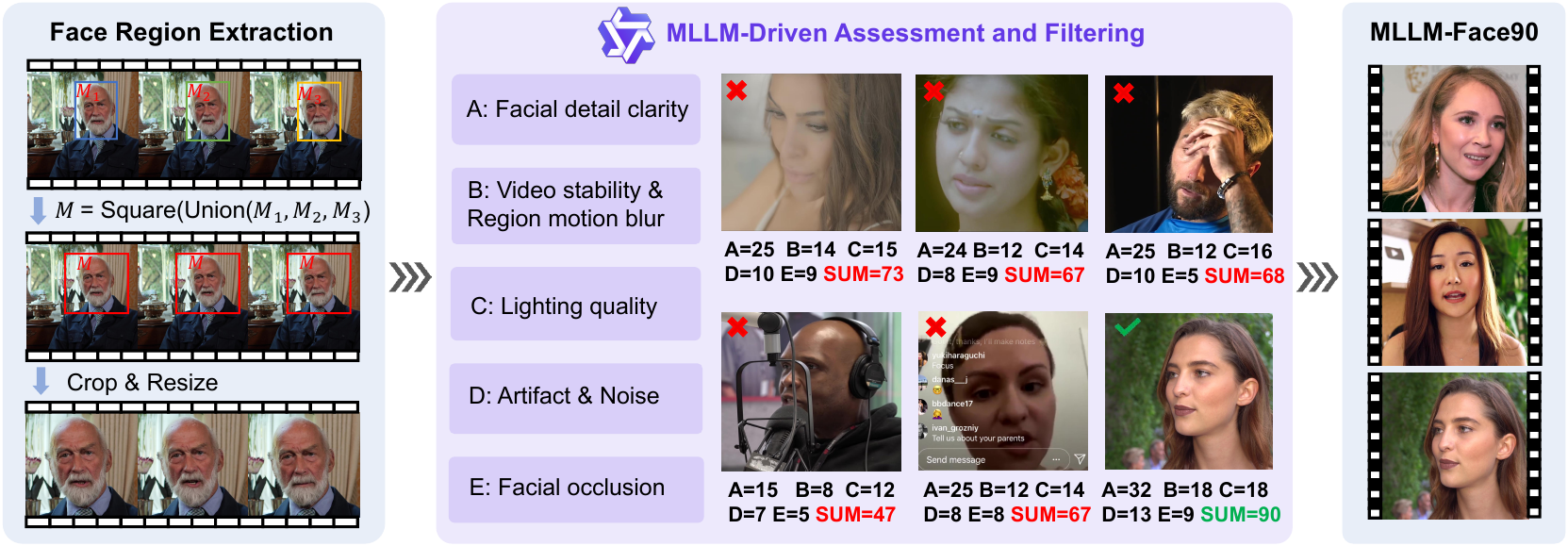} 
    \caption{Pipeline of the proposed MLLM-driven high-quality face video filtering. First, face regions are extracted and cropped to facilitate the model's focus on facial features. Next, a meticulously designed set of visual quality assessment prompts is utilized to evaluate each video from multiple quality perspectives using the powerful Qwen2.5-VL.}
    \label{fig:mllm}
    \vspace{3mm}
\end{figure*}

\subsection{MLLM-Driven High-Quality Video Filtering}
\label{sec:dataset}
Existing face-centric datasets such as VFHQ suffer from two critical limitations:
(1) facial regions occupy only a small portion of each frame, causing models to overfit to background reconstruction rather than learning fine-grained facial texture representations;
(2) many video exhibit severe degradations, including motion blur, poor illumination, and occlusions. 
The quality of training data fundamentally determines the upper bound of model performance. Training on such degraded samples leads to outputs with unrealistic textures, as models cannot reproduce what they have never seen in high quality.

Based on these observations, we propose a novel MLLM-driven high-quality video filtering pipeline. This pipeline produces a curated, face-centric dataset called \textbf{MLLM-Face90}, as illustrated in Fig.~\ref{fig:mllm}.
% Our approach consists of face region extraction followed by MLLM-based quality assessment.

\noindent\textbf{Face Region Extraction.} 
We employ a robust face parsing model~\citep{yu2018bisenet} to segment each frame and extract tight bounding boxes around facial regions. To maintain spatial consistency across the video sequence, we aggregate all per-frame bounding boxes into a unified crop window that encompasses the face throughout all frames. This global window is then adjusted to a square shape to prevent facial distortion during subsequent resizing to square training dimensions. Finally, each frame is cropped according to this unified bounding box and resized to the target training dimensions. This procedure effectively eliminates background interference while preserving facial geometry integrity for subsequent quality assessment and model training.

\noindent\textbf{MLLM-Driven Quality Assessment and Filtering.} 
To ensure high-quality training data, we design an automated quality assessment pipeline using Qwen2.5-VL~\citep{Qwen2.5-VL} with a meticulously crafted evaluation protocol specifically tailored for face restoration. Our protocol evaluates five critical dimensions: (1) facial detail clarity, (2) video stability and regional motion blur, (3) lighting quality, (4) compression artifacts and noise, and (5) facial occlusion. Each dimension is scored following strict rubrics, with bonus or penalty adjustments for outstanding quality or evident deficiencies. The protocol explicitly prioritizes critical facial regions (eyes, mouth, teeth, nose), penalizing motion blur or degradation in these areas while requiring natural lighting and artifact-free quality (see complete prompt in Supplementary Material Section~\ref{app:mllm_prompt}).
Only videos achieving an overall quality score above 90 out of 100 are retained, forming our curated benchmark dataset, \textbf{MLLM-Face90}. Fine-tuning on MLLM-Face90 leads to substantial performance improvements, as shown in Table~\ref{table:training_dataset_ablation}.

\begin{table*}[ht]  % VFHQ-test，占两栏宽度
\centering
\setlength{\tabcolsep}{7pt} 
\caption{Quantitative comparison on the \textbf{VFHQ-test} dataset. The best and second-best methods are highlighted in \textcolor{red}{red} and \textcolor{blue}{blue} respectively. VividFace achieves superior performance across all metrics.}
\label{table: vfhq-test}
\begin{tabular}{lccccccccc}
\toprule
\multirow{2}{*}{Method} & \multicolumn{3}{c}{Quality and Fidelity} & \multicolumn{3}{c}{Pose Consistency} & \multicolumn{2}{c}{Temporal Consistency} \\
\cmidrule(lr){2-4} \cmidrule(lr){5-7} \cmidrule(lr){8-9}
 & PSNR↑ & SSIM↑ & LPIPS↓ & IDS↑ & AKD*↓ & FaceCons↑ & FasterVQA↑ & FVD↓ \\
\midrule
BasicVSR++\cite{chan2022basicvsr++} & 25.64 & 0.7860 & 0.3902 & 0.7796 & 9.2778 & 0.7475 & 0.2108 & 1073.70 \\
RealBasicVSR\cite{chan2022realbasicvsr} & 26.82 & 0.7793 & 0.2656 & 0.8046 & 6.3262 & 0.7442 & 0.8086 & 308.90 \\
MGLD-VSR\cite{yang2023mgldvsr} & 27.37 & 0.8111 & 0.2151 & 0.8350 & 4.6829 & 0.7297 & 0.7977 & 285.30 \\
STAR\cite{xie2025star} & 24.66 & 0.7729 & 0.3393 & 0.8077 & 6.6215 & 0.7605 & 0.6643 & 464.51 \\
SeedVR-3B\cite{wang2025seedvr} & 27.04 & 0.7860 & 0.2271 & 0.8523 & 4.4892 & 0.7905 & 0.8025 & 126.14 \\
SeedVR2-3B\cite{wang2025seedvr2} & 27.75 & 0.8420 & \textcolor{blue}{0.1538} & \textcolor{blue}{0.8887} & 3.9975 & 0.8013 & 0.8194 & 116.56 \\
\midrule
PGTFormer\cite{xu2024pgtformer} & \textcolor{blue}{28.78} & \textcolor{blue}{0.8460} & 0.1837 & 0.8612 & 4.3572 & 0.7298 & \textcolor{blue}{0.8484} & 197.02 \\
BFVR-STC\cite{wang2025bfvr} & 24.37 & 0.7858 & 0.3383 & 0.7793 & 8.2443 & 0.7179 & 0.4594 & 700.20 \\
KEEP\cite{feng2024keep} & 27.50 & 0.8152 & 0.2376 & 0.7950 & 4.5966 & 0.7529 & 0.7986 & 388.60 \\
SVFR\cite{wang2025svfr} & 28.09 & 0.8304 & 0.1578 & 0.8641 & 4.0932 & \textcolor{blue}{0.8025} & 0.8404 & \textcolor{blue}{103.32} \\
DicFace\cite{chen2025dicface} & 28.25 & 0.8313 & 0.2424 & 0.8854 & \textcolor{blue}{3.9682} & 0.7634 & 0.7207 & 340.76\\
\midrule
\textbf{VividFace (Ours)} & \textcolor{red}{30.03} & \textcolor{red}{0.8534} & \textcolor{red}{0.1112} & \textcolor{red}{0.9128} & \textcolor{red}{3.5319} & \textcolor{red}{0.8111} & \textcolor{red}{0.8855} & \textcolor{red}{79.14} \\
\bottomrule
\end{tabular}%
\vspace{2mm}
\end{table*}

\section{Experiments} \label{sec: exp}
\subsection{Implementation details}
\label{ssec: implementation}
\textbf{Training Details.} We adopt a coarse-to-fine strategy: first, 3,000 clips randomly sampled from VFHQ~\citep{xie2022vfhq} are used for coarse training, followed by fine-tuning on MLLM-Face90, a curated set of 1,957 high-quality clips. To simulate real-world degradations, we follow~\citep{feng2024keep} to synthesize low-quality data: $y = \left[ (x \circledast k_\sigma) \downarrow_r + n_\delta \right]_{\text{FFMPEG}_{\text{crf}}}$, where $x$, $y$ are high- and low-quality videos. $\circledast$ represents convolution, $k_\sigma$ and $n_\delta$ are the Gaussian blur kernel and noise, and $\downarrow_r$ indicates $r\times$ downsampling. During synthesis, $\sigma$, $r$, $\delta$, and $\text{crf}$ are randomly sampled from $[0.1, 10]$, $[1, 4]$, $[0, 10]$, and $[18, 25]$, respectively. The hyperparameters $\lambda$ and $p$ are empirically set to 0.1 and 0.5, respectively. The frame resolutions for latent and pixel training stages are $81 \times 512 \times 512$ and $13 \times 512 \times 512$. All experiments are conducted on eight NVIDIA A100 GPUs, with a batch size of 32 and a learning rate of $1 \times 10^{-4}$ for a total of 32,000 iterations.

\noindent\textbf{Evaluation.} To comprehensively validate VividFace, we test on both synthetic and real-world benchmarks. For synthetic evaluation, following previous work, we use the official VFHQ-test dataset~\citep{xie2022vfhq} with the aforementioned degradation model, containing 50 high-quality video clips. For real-world evaluation, we follow prior protocols using RFV-LQ~\citep{wang2024rfvlq}. RFV-LQ contains 329 low-quality face videos meticulously curated from diverse real-world sources, including old talk shows, TV series, and movies, providing a robust testbed for evaluating the method's robustness across various real-world conditions.

\noindent\textbf{Metrics.} To facilitate a comprehensive and rigorous evaluation, we employ a diverse set of video quality assessment metrics spanning multiple dimensions of model performance. Specifically, we assess results from three key perspectives: Quality and Fidelity, Pose Consistency, and Temporal Consistency. For Quality and Fidelity, we use six representative metrics: PSNR, SSIM~\citep{wang2004psnrssim}, and LPIPS~\citep{zhang2018lpips} (reference-based), as well as NIQE~\citep{mittal2012niqe}, MUSIQ~\citep{ke2021musiq}, and CLIP-IQA~\citep{wang2023clipiqa} (no-reference). To measure Pose Consistency, we adopt IDS, AKD, and FaceCons~\citep{DBLP:conf/eccv/FengLL24}, as well as TLME~\citep{DBLP:conf/ijcai/Xu00YL24}. We multiplie AKD and TLME by 1000, denoted as AKD* and TLME*, respectively. For Temporal Consistency, we employ FasterVQA~\citep{wu2023fastervqa} and FVD~\citep{DBLP:conf/iclr/UnterthinerSKMM19}. 
% This diverse set of metrics enables robust and systematic evaluation of video face enhancement.

\begin{table*}[ht]  % RFV-LQ，占两栏宽度
\centering
\setlength{\tabcolsep}{9pt} 
\renewcommand{\arraystretch}{1.0} 
\caption{Quantitative comparison on the \textbf{real-world} RFV-LQ dataset. The best and second-best methods are highlighted in \textcolor{red}{red} and \textcolor{blue}{blue} respectively. Our method demonstrates superior performance in real-world scenarios.}
\label{table: rfv-lq}
\begin{tabular}{lccccccc}
\toprule
\multirow{2}{*}{Method} & \multicolumn{3}{c}{Quality and Fidelity} & \multicolumn{2}{c}{Pose Consistency} & \multicolumn{1}{c}{Temporal Consistency} \\
\cmidrule(lr){2-4} \cmidrule(lr){5-6} \cmidrule(lr){7-7}
 & NIQE↓ & MUSIQ↑ & CLIP-IQA↑ & TLME*↓ & FaceCons↑ & FasterVQA↑ \\
\midrule
BasicVSR++\cite{chan2022basicvsr++} & 6.2983 & 30.8569 & 0.2005 & 7.3107 & 0.7207 & 0.2651 \\
RealBasicVSR\cite{chan2022realbasicvsr} & \textcolor{red}{5.0402} & 63.1429 & 0.5407 & 6.5338 & 0.7392 & 0.7305\\
MGLD-VSR\cite{yang2023mgldvsr} & 5.9269 & 62.7775 & 0.5593 & 6.2764 & 0.7359 & 0.7630 \\
STAR & 5.5227 & \textcolor{blue}{64.4846} & \textcolor{blue}{0.5416} & 6.4900 & 0.7222 & 0.7657\\
SeedVR-3B\cite{wang2025seedvr} & 5.3900 & 52.8371 & 0.4759 & 6.6668 & 0.7430 & 0.6649 \\
SeedVR2-3B\cite{wang2025seedvr2} & 6.6548 & 54.4296 & 0.4051 & 6.5437 & \textcolor{blue}{0.7593} & 0.6059 \\
\midrule
PGTFormer\cite{xu2024pgtformer} & 6.7676 & 59.5214 & 0.4709 & 6.2961 & 0.7151 & \textcolor{blue}{0.7744} \\
BFVR-STC\cite{wang2025bfvr} & 6.8477 & 45.5984 & 0.3372 & 6.7475 & 0.6928 & 0.5149 \\
KEEP\cite{feng2024keep} & 6.2016 & 60.9558 & 0.5054 & \textcolor{blue}{6.1623} & 0.7392 & 0.7255 \\
SVFR\cite{wang2025svfr} & 7.0772 & 54.2877 & 0.3907 & 6.1478 & 0.7527 & 0.6286\\
DicFace\cite{chen2025dicface} & 6.8448 & 53.9808 & 0.4525 & 6.2385 & 0.7413 & 0.6185 \\
\midrule
\textbf{VividFace (Ours)} & \textcolor{blue}{5.1987} & \textcolor{red}{64.4911} & \textcolor{red}{0.5678} & \textcolor{red}{6.0064} & \textcolor{red}{0.7665} & \textcolor{red}{0.8227}\\
\bottomrule
\end{tabular}%
% \vspace{-2mm}
\end{table*}

\noindent\textbf{Compared Methods.} We evaluate our approach against three distinct categories of classic and representative compared methods. First, we select widely-used Video Super-Resolution (VSR) models, including BasicVSR++~\citep{chan2022basicvsr++}, RealBasicVSR~\citep{chan2022realbasicvsr}, MGLD-VSR~\citep{yang2023mgldvsr}, STAR~\citep{xie2025star}, SeedVR-3B~\citep{wang2025seedvr}, and SeedVR2-3B~\citep{wang2025seedvr2}.
% , which enhance overall video quality but lack specialized facial restoration. 
Second, we include established Face Image-based Restoration (FIR) models, such as CodeFormer~\citep{zhou2022codeformer} and DifFace~\citep{yue2024difface}, which restore facial details independently for each frame. Third, we compare with state-of-the-art Face Video Restoration (FVR) models including PGTFormer~\citep{xu2024pgtformer}, BFVR-STC~\citep{wang2025bfvr}, KEEP~\citep{feng2024keep}, SVFR~\citep{wang2025svfr}, and DicFace~\citep{chen2025dicface}. For all experiments, we use the same degradation settings and official implementations. For methods restricted to processing aligned facial regions, including CodeFormer, DifFace, KEEP, and DicFace, we adopt a unified approach following the KEEP pipeline. The background is first enhanced using ESRGAN~\citep{wang2021realeargan}, and then composited back into the original frames.

\subsection{Comparison with State-of-the-Art}
\label{ssec:sota}

\textbf{Quantitative Results.}
We present a comprehensive performance comparison on the VFHQ-test benchmark in Table~\ref{table: vfhq-test} and the real-world RFV-LQ benchmark in Table~\ref{table: rfv-lq}. VividFace demonstrates outstanding results, consistently surpassing both VSR and FVR methods across all evaluation metrics. Specifically, in Table~\ref{table: vfhq-test}, VividFace significantly outperforms previous methods in PSNR, LPIPS, and FVD, achieving consistent superiority across multiple metrics. In Table~\ref{table: rfv-lq}, diverse no-reference video quality assessment metrics further confirm its robustness and superior visual quality in real-world scenarios.
Additional comparisons with FIR methods are provided in the Supplementary Material (Table~\ref{table:appendix_fir}) show VividFace achieves notably higher temporal consistency, highlighting the effectiveness of our approach.

% \noindent\textbf{Qualitative Results.} Comprehensive qualitative results on the VFHQ-test and real-world RFV-LQ datasets are presented in Figure~\ref{figure:qualitive_comparison_1} and Figure~\ref{figure:qualitive_comparison_2}. Existing methods often yield blurred or over-smoothed results and fail to reconstruct realistic, identity-consistent structures across key regions such as the eyes, eyebrows, and mouth. In contrast, VividFace demonstrates superior capability in recovering more realistic and visually pleasing results: it faithfully restores fine-grained features while preserving surrounding facial structures, maintaining consistency with the ground-truth identity, and avoiding the over-smoothing artifacts observed in competing approaches. Additional qualitative results are provided in the \textbf{supplementary  video files}.

\begin{figure*}[ht]
    \centering
    \includegraphics[width=\linewidth]{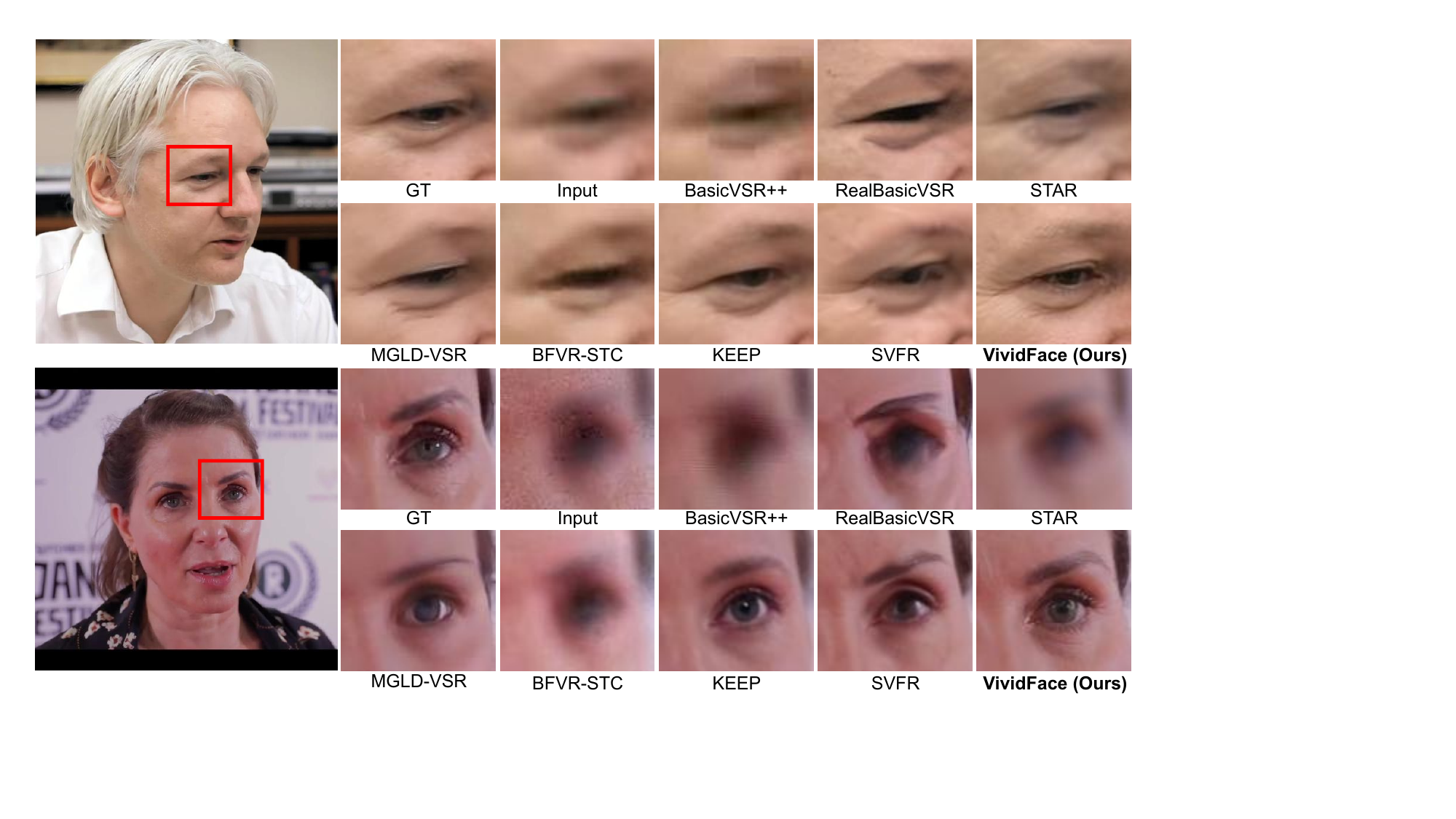}
% \vspace{-2em}
\caption{Visual comparison with existing methods on VFHQ-test. VividFace exhibits more realistic and visually pleasing facial details, and produces results that are closer to the ground truth.}
\label{figure:qualitive_comparison_1}
\vspace{-4mm}
\end{figure*}

\begin{figure*}[ht]
    \centering
    \includegraphics[width=\linewidth]{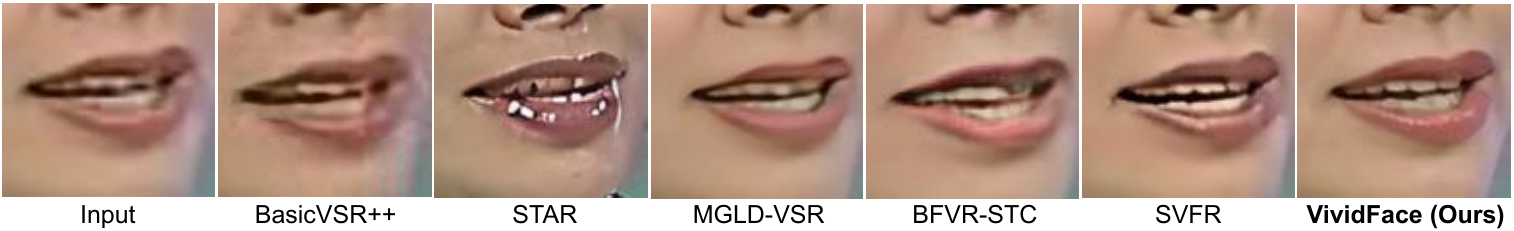}
% \vspace{-2em}
\caption{Qualitative comparison on real-world RFV-LQ dataset. The results highlight VividFace's strong capability to address complex real-world degradations.}
\label{figure:qualitive_comparison_2}
% \vspace{2mm}
\end{figure*}

\noindent\textbf{Qualitative Results.} 
% Comprehensive qualitative results on the VFHQ-test and real-world RFV-LQ datasets are shown in Figure~\ref{figure:qualitive_comparison_1} and Figure~\ref{figure:qualitive_comparison_2}. 
% Existing methods often produce blurred or over-smoothed outputs, struggling to reconstruct realistic structures in key regions such as the eyes, eyebrows, and mouth. In contrast, VividFace effectively restores fine-grained details, preserves identity consistency, and avoids over-smoothing artifacts. Additional qualitative results are provided in the \textbf{supplementary  video files}.
Comprehensive qualitative results on the VFHQ-test and real-world RFV-LQ datasets are presented in Figure~\ref{figure:qualitive_comparison_1} and Figure~\ref{figure:qualitive_comparison_2}. 
% Existing methods often produce blurred or over-smoothed details and fail to reconstruct realistic, identity-consistent structures across key regions such as the eyes, eyebrows, and mouth. 
Existing methods often produce blurred or over-smoothed outputs, struggling to reconstruct realistic structures in key regions such as the eyes, eyebrows, and mouth. 
In contrast, VividFace demonstrates superior capability in recovering more realistic and visually pleasing results: it faithfully restores fine-grained features while preserving surrounding facial structures, maintaining consistency with the ground-truth identity, and avoiding the over-smoothing artifacts observed in competing approaches. Additional qualitative results are provided in the \textbf{supplementary video files}.

\noindent\textbf{Running Time Comparisons.} Table~\ref{table:running_time} compares the inference speed and performance of several leading methods, all evaluated on identical hardware (single 80GB A100 GPU) using 81-frame 512×512 videos. Notably, VividFace completes inference in just \textbf{7.43 seconds}, being approximately \textbf{12$\times$ faster} than SVFR and \textbf{2.7$\times$ faster} than SeedVR2-3B, the closest one-step competitor. Importantly, VividFace achieves this speed without compromising perceptual quality or identity preservation, as evidenced by its superior LPIPS and IDS scores. These results highlight VividFace's clear advantage in both efficiency and output fidelity.

\begin{table}[ht]
\centering
\small
\caption{Running time comparison on 81-frame 512$\times$512 videos across various methods. VividFace achieves the fastest inference speed while consistently delivering superior visual quality and better identity preservation compared to other approaches.}
% \caption{Running time comparison on 81-frame 512$\times$512 videos. VividFace achieves the fastest speed with high visual quality.}
% \vspace{5pt} 
\label{table:running_time}
\begin{tabular}{lcccc}
\toprule
Method & Step & Time (s)$\downarrow$ & LPIPS$\downarrow$ & IDS$\uparrow$ \\
\midrule
MGLD-VSR\cite{yang2023mgldvsr} & 50 & 98.73 & 0.215 & 0.835 \\
STAR\cite{xie2025star} & 15 & 456.83 & 0.339 & 0.808 \\
SeedVR-3B\cite{wang2025seedvr} & 50 & 151.43 & 0.227 & 0.852 \\
SeedVR2-3B\cite{wang2025seedvr2} & 1 & 19.98 & 0.153 & 0.888 \\
SVFR\cite{wang2025svfr} & 30 & 94.60 & 0.157 & 0.864 \\
\midrule
\textbf{VividFace (Ours)} & 1 & \textbf{7.43} & \textbf{0.113} & \textbf{0.912} \\
\bottomrule
\end{tabular} 
\vspace{-1.5mm}
\end{table}
% \vspace{3mm}

\subsection{Ablation Studies}
\label{ssec: ablation}

\textbf{Effect of Face-Focused Training Probability $p$.}
We investigate the impact of face-focused training probability $p$ in Eq.~\ref{eq:stochastic_loss}, which controls the balance between face-focused and global reconstruction objectives. As shown in Table~\ref{table:mask_probability}, we evaluate three different values: 0 (pure global training), 0.5 (balanced stochastic training), and 1 (pure face-focused training). The results demonstrate that $p=0.5$ achieves optimal performance. Pure global training leads to suboptimal facial detail recovery, while pure face-focused training compromises overall video quality due to insufficient global context learning. The balanced approach effectively combines both objectives.

\begin{table}[ht]
  \centering
  % \caption{Effect of face-focused training probability $p$.}
  \caption{Effect of face-focused training probability $p$. The balanced probability $p=0.5$ achieves best performance through joint face-focused and global reconstruction.}
  \setlength{\tabcolsep}{24pt}
  \begin{tabular}{lcc}
    \toprule
    $p$ & LPIPS$\downarrow$ & IDS$\uparrow$ \\
    \midrule
    0  &  0.1229 &  0.9073 \\
    \textbf{0.5} &  \textbf{0.1112} &  \textbf{0.9128} \\
    1  & 0.1309 & 0.9025 \\
    \bottomrule
  \end{tabular}
  \label{table:mask_probability}
  \vspace{-2mm}
\end{table}

\noindent\textbf{Effectiveness of Joint Latent-Pixel Training Strategy.}
We validate our Joint Latent-Pixel Face-Focused Training strategy by progressively adding training components. Table~\ref{table:latent_rgb} presents three configurations: baseline without face-focused training, latent space only, and joint training in both spaces. Adding latent space training alone slightly degrades performance, suggesting that latent-only optimization is insufficient for fine-grained facial detail recovery. However, the complete joint training strategy significantly improves both metrics, demonstrating that multi-space optimization is crucial for optimal facial enhancement quality.

% 第二个表格
\begin{table}[ht]
  \centering
  % \caption{Ablation study on face-focused training strategy.}
  \caption{Ablation study on Joint Latent-Pixel Face-Focused Training. The joint training strategy clearly outperforms individual components, validating the necessity of multi-space optimization.}
  \vspace{0mm}
  \setlength{\tabcolsep}{15pt}
  \begin{tabular}{lccc}
    \toprule
    Latent & Pixel & LPIPS$\downarrow$ & IDS$\uparrow$\\
    \midrule
    $\times$ & $\times$ & 0.1229 & 0.9073 \\
    $\checkmark$ & $\times$ & 0.1261 & 0.9050\\
    $\checkmark$ & $\checkmark$ & \textbf{0.1112} &  \textbf{0.9128} \\
    \bottomrule
  \end{tabular}
  \label{table:latent_rgb}
\end{table}

\noindent\textbf{Impact of MLLM-Face90 High-Quality Dataset.}
We evaluate the contribution of our MLLM-Face90 dataset by comparing models trained on different data configurations. Table~\ref{table:training_dataset_ablation} compares using only VFHQ-3K versus incorporating our curated dataset. The incorporation of MLLM-Face90 leads to clear improvements across both metrics, validating the effectiveness of our MLLM-driven data curation pipeline in providing high-quality face-centric training data for learning authentic facial textures.

\begin{table}[t]
  \centering
  % \caption{Ablation study on model performance with our proposed MLLM-Face90 dataset.}
  \caption{Ablation study on the MLLM-Face90 dataset. Incorporating our high-quality curated dataset improves performance.}
  % \vspace{1mm}
  \setlength{\tabcolsep}{16pt} % 调整间距
  \begin{tabular}{lcc}
    \toprule
    Dataset & LPIPS$\downarrow$ & IDS$\uparrow$ \\
    \midrule
    VFHQ-3K &  0.1285 &  0.9046 \\
    \textbf{+MLLM-Face90} & \textbf{0.1112} & \textbf{0.9128} \\
    \bottomrule
  \end{tabular}
  \label{table:training_dataset_ablation}
  \vspace{-3mm}
\end{table}

\section{Conclusion} 
\label{sec:conclusion}
In this work, we present VividFace, an efficient one-step diffusion framework that advances video face enhancement through three key innovations. First, by reformulating multi-step diffusion into single-step flow matching built upon the text-to-video model WANX, we achieve substantial efficiency gains without compromising visual fidelity. Second, our Joint Latent-Pixel Face-Focused Training strategy leverages spatiotemporally aligned facial masks to enable targeted optimization of critical facial regions across both representation spaces. Third, our MLLM-driven data curation pipeline facilitates automated construction of high-quality face video datasets, enhancing model generalization and robustness. Comprehensive evaluations validate the effectiveness of each component, with VividFace consistently outperforming existing methods across multiple benchmarks. We hope our released resources encourage further research on efficient generative video restoration.

% In this work, we present VividFace, a novel and efficient one-step diffusion framework for video face enhancement. By leveraging the powerful WANX video model as the backbone, our approach benefits from robust spatio-temporal representations, enabling more accurate and consistent restoration of facial details across frames. The integration of our Joint Latent-Pixel Face-Focused Training strategy with stochastic switching between facial and global optimization objectives within a single-step flow matching paradigm significantly accelerates inference while preserving high perceptual quality. Furthermore, our MLLM-driven data curation pipeline facilitates automated construction of high-quality face video datasets, further enhancing model generalization and robustness. 
% Extensive experiments demonstrate that VividFace achieves state-of-the-art performance in perceptual quality, identity preservation, and temporal stability.
% We believe our work demonstrates the potential of combining advanced video backbones with efficient generative frameworks for high-fidelity video face restoration and provides useful resources for future research.
{
    \small
    \bibliographystyle{ieeenat_fullname}
    \bibliography{main}
}

\clearpage
\setcounter{page}{1}
\maketitlesupplementary

% \balance

% \section{Rationale}
% \label{sec:rationale}
% % 
% Having the supplementary compiled together with the main paper means that:
% % 
% \begin{itemize}
% \item The supplementary can back-reference sections of the main paper, for example, we can refer to \cref{sec:intro};
% \item The main paper can forward reference sub-sections within the supplementary explicitly (e.g. referring to a particular experiment); 
% \item When submitted to arXiv, the supplementary will already included at the end of the paper.
% \end{itemize}
% % 
% To split the supplementary pages from the main paper, you can use \href{https://support.apple.com/en-ca/guide/preview/prvw11793/mac#:~:text=Delete%20a%20page%20from%20a,or%20choose%20Edit%20%3E%20Delete).}{Preview (on macOS)}, \href{https://www.adobe.com/acrobat/how-to/delete-pages-from-pdf.html#:~:text=Choose%20%E2%80%9CTools%E2%80%9D%20%3E%20%E2%80%9COrganize,or%20pages%20from%20the%20file.}{Adobe Acrobat} (on all OSs), as well as \href{https://superuser.com/questions/517986/is-it-possible-to-delete-some-pages-of-a-pdf-document}{command line tools}.

\appendix

\section{More visual and performance comparison}

\begin{figure*}[hb]
    \centering
    \includegraphics[width=\linewidth]{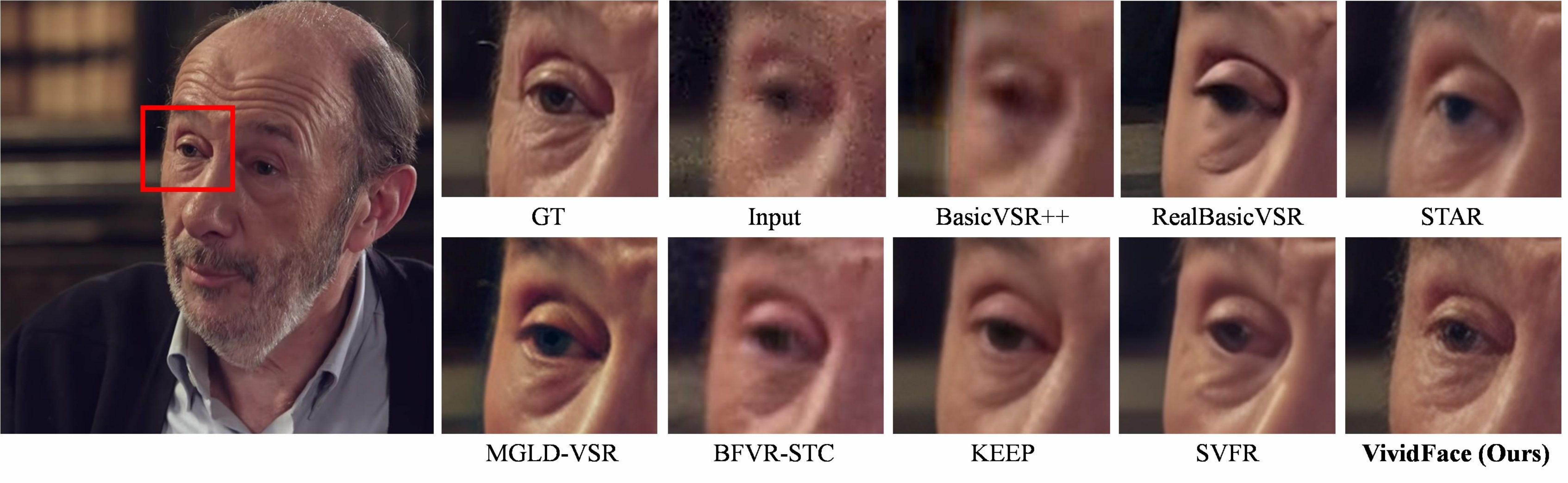}
\vspace{-1em}
\caption{More qualitative comparison on VFHQ-test dataset. VividFace effectively recovers subtle skin details and enhances perceptual quality.}
\label{figure:appendix_qualitive_2}
\vspace{-1em}
\end{figure*}

\paragraph{More qualitative comparisons with face restoration methods.} Figure~\ref{figure:appendix_qualitive_2} presents additional qualitative comparisons on the VFHQ-test dataset. As illustrated, our VividFace faithfully recovers skin details such as wrinkles and texture around the eye region. These results further validate the capability of VividFace to achieve photorealistic facial enhancement.

\paragraph{More performance comparisons with face restoration methods.} 
As shown in Table~\ref{table:appendix_fir}, FIR methods exhibit substantially inferior performance compared to our video-native approach across all evaluation dimensions. Notably, while FIR methods achieve reasonable per-frame quality metrics, they suffer from severe degradation in pose consistency and catastrophic failure in temporal coherence. DifFace, despite being a recent diffusion-based method, shows particularly poor identity preservation and facial geometry consistency, suggesting that frame-independent stochastic generation fundamentally lacks the temporal modeling required for video restoration. These results demonstrate the critical importance of video-aware processing and validate the exceptional capability of our approach in maintaining spatio-temporal consistency for face video restoration.

\begin{table*}[hb]
\centering
\caption{Comparisons of face image restoration (FIR) methods on the VFHQ-test dataset.}
\label{table:appendix_fir}
\begin{tabular}{ccccccccccc} 
\toprule
\multicolumn{2}{c}{\multirow{2}{*}[-0.3em]{Method}} & \multicolumn{3}{c}{Quality and Fidelity} 
& \multicolumn{3}{c}{Pose Consistency} 
& \multicolumn{2}{c}{Temporal Consistency} 
\\
\cmidrule(lr){3-5}  \cmidrule(lr){6-8}  \cmidrule(lr){9-10} 
 & & PSNR↑ & SSIM↑ & LPIPS↓ & IDS↑ & AKD*↓ & FaceCons↑ & FasterVQA↑ & FVD↓ \\
\midrule
\multirow{2}{*}{FIR} & CodeFormer~\citep{zhou2022codeformer} & 27.27 & 0.8023 & 0.2391 & 0.7719 & 5.5370 & 0.7156 &  0.8625 & 331.99 \\
& DifFace~\citep{yue2024difface} & 26.73 & 0.7919 & 0.2538 & 0.5998 & 7.6810 & 0.4544 & 0.8278 &  904.14\\
\midrule
\multicolumn{2}{l}{\textbf{VividFace (Ours)}} & \textcolor{red}{30.03} & \textcolor{red}{0.8534} & \textcolor{red}{0.1112} & \textcolor{red}{0.9128} & \textcolor{red}{3.5319} & \textcolor{red}{0.8111} & \textcolor{red}{0.8855} & \textcolor{red}{79.14} \\
\bottomrule
\end{tabular}
\vspace{-2em}
\end{table*}

\label{appendix: experimental}

\begin{figure}[ht]
    \centering
    \includegraphics[width=\linewidth]{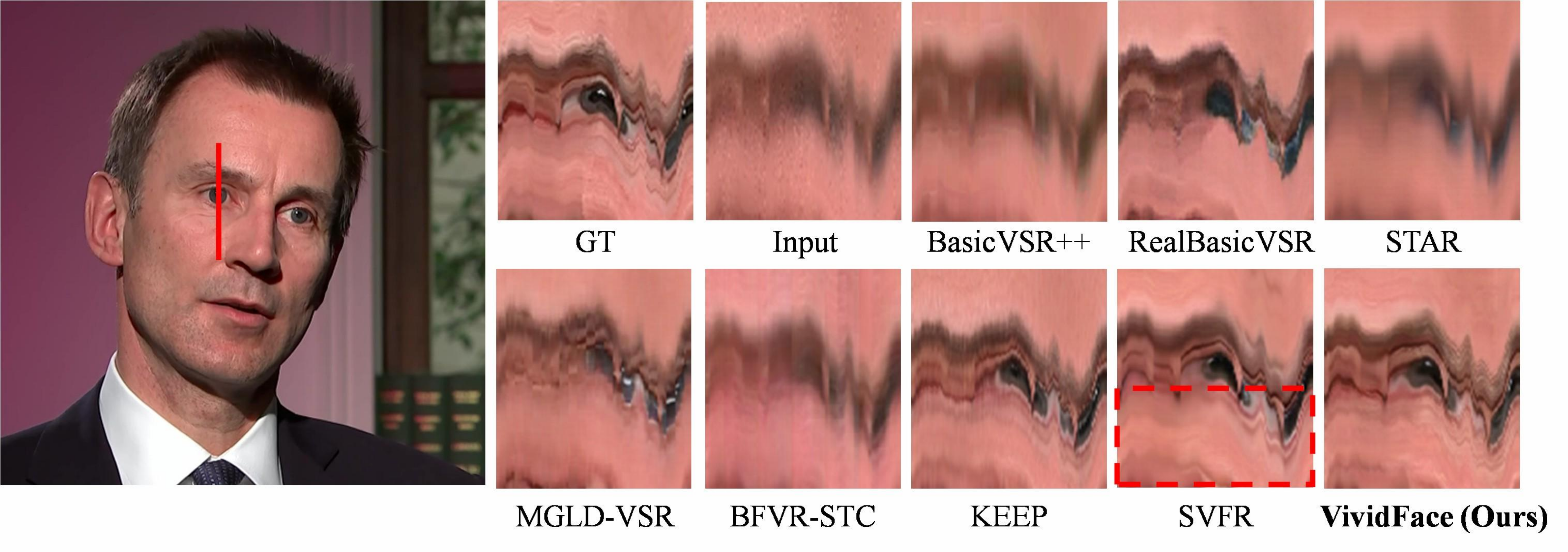}
% \vspace{-2em}
\caption{Visualization of temporal consistency comparison. VividFace achieves temporal results that are closer to the ground truth, verifying its stronger temporal consistency.}

\label{figure:appendix_temporal}
\end{figure}

\paragraph{Visualization of temporal consistency comparison.} Temporal consistency is a crucial aspect of video enhancement. Therefore, we further compare the temporal consistency performance of existing methods, as shown in Figure~\ref{figure:appendix_temporal}. Specifically, we select the region marked by the red line and display its continuous representations across different frames. It can be observed that VividFace demonstrates superior temporal consistency with significantly reduced jitter compared to other approaches, and its results are much closer to the ground truth.

\section{MLLM Prompt for Video Quality Assessment}
\label{app:mllm_prompt}

In this section, we present the detailed prompt used for multi-dimensional video quality assessment. The prompt evaluates videos across five key dimensions: \textbf{facial detail clarity}, \textbf{video stability and motion blur}, \textbf{lighting quality}, \textbf{artifact and noise level}, and \textbf{facial occlusion}. It employs a comprehensive 100-point scoring system with bonus and penalty mechanisms to ensure rigorous selection of premium training data.

\clearpage

% 切换到单栏模式，让 prompt 占据全宽
\onecolumn

\begin{tcolorbox}[
    colback=gray!5,
    colframe=gray!75!black,
    title=Premium Face Video Quality Evaluation Prompt,
    fonttitle=\bfseries,
    boxrule=0.5pt,
    arc=2mm,
    breakable]
\small\ttfamily
You are an expert face video quality inspector specializing in premium face restoration training data. Evaluate this video with STRICT criteria to identify only the highest quality samples suitable for premium model training.

\vspace{0.2cm}
\textbf{Evaluation Criteria (Total: 100 points) - STRICT GRADING:}

\vspace{0.1cm}
1. \textbf{Facial Detail Clarity (35 points)}
\begin{itemize}[leftmargin=1em, itemsep=0cm, topsep=0.1cm]
   \item 0-12: Severely degraded, facial features barely distinguishable
   \item 13-21: Moderate quality, basic features visible but lacking fine details
   \item 22-28: Good quality with visible skin texture and facial features, BUT penalize if key regions (eyes, mouth, teeth) show motion blur
   \item 29-32: Excellent clarity with clear pores, fine lines, and detailed texture across ALL facial regions
   \item 33-35: Perfect clarity with crisp micro-details (individual eyelashes, teeth edges, lip texture clearly visible)
\end{itemize}

\vspace{0.1cm}
2. \textbf{Video Stability \& Regional Motion Blur (20 points)}
\begin{itemize}[leftmargin=1em, itemsep=0cm, topsep=0.1cm]
   \item 0-6: Severe motion blur or instability affecting entire face
   \item 7-11: Noticeable camera shake OR significant motion blur in key facial regions (eyes, mouth, teeth)
   \item 12-15: Minor overall stability issues, but critical facial features remain sharp
   \item 16-18: Very stable with minimal motion blur, all key facial regions clear
   \item 19-20: Perfect stability across all frames, no motion blur in any facial region
\end{itemize}

\vspace{0.1cm}
3. \textbf{Lighting Quality (20 points)}
\begin{itemize}[leftmargin=1em, itemsep=0cm, topsep=0.1cm]
   \item 0-6: Extreme lighting conditions that obscure facial features
   \item 7-11: Acceptable lighting with noticeable issues (uneven shadows, slight over/under exposure)
   \item 12-15: Good lighting with minor imperfections
   \item 16-18: Excellent natural lighting with proper facial modeling
   \item 19-20: Perfect studio-quality lighting with optimal facial structure revelation
\end{itemize}

\vspace{0.1cm}
4. \textbf{Artifact \& Noise Level (15 points)}
\begin{itemize}[leftmargin=1em, itemsep=0cm, topsep=0.1cm]
   \item 0-4: Heavy compression artifacts, noise, or digital distortions
   \item 5-7: Noticeable artifacts that affect facial details
   \item 8-10: Minor artifacts present but don't significantly impact quality
   \item 11-13: Minimal artifacts, high video quality
   \item 14-15: No visible artifacts, pristine video quality
\end{itemize}

\vspace{0.1cm}
5. \textbf{Facial Occlusion (10 points)}
\begin{itemize}[leftmargin=1em, itemsep=0cm, topsep=0.1cm]
   \item 0-2: Significant occlusion (>25\% of face covered by objects, hands)
   \item 3-4: Moderate occlusion (10-25\% covered)
   \item 5-6: Minor occlusion (5-10\% covered), most facial features visible
   \item 7-8: Minimal occlusion (<5\% covered), all key facial features clearly visible
   \item 9-10: No occlusion, complete facial visibility
\end{itemize}

\vspace{0.2cm}
\textbf{Critical Facial Regions Check:}
\begin{itemize}[leftmargin=0.5em, itemsep=0cm, topsep=0.1cm]
\item Eyes: Must be sharp with visible iris details, eyelashes clearly defined
\item Mouth/Lips: Lip texture and edges must be crisp, no blur during speech
\item Teeth: Individual teeth edges must be clearly visible when shown
\item Nose: Nostril details and nose bridge must be sharp
\end{itemize}

\vspace{0.2cm}
\textbf{STRICT QUALITY THRESHOLDS:}
\begin{itemize}[leftmargin=0.5em, itemsep=0cm, topsep=0.1cm]
\item Premium Training Data: Score >= 85 (Top 10-15\% of videos)
\item High-Quality Training Data: Score >= 80 (Top 20-25\% of videos)
\item Standard Training Data: Score >= 75 (Top 40\% of videos)
\item Below Standard: Score < 75 (Consider discarding for premium training)
\end{itemize}

\vspace{0.2cm}
\textbf{Additional Quality Factors (Bonus/Penalty):}
\begin{itemize}[leftmargin=0.5em, itemsep=0cm, topsep=0.1cm]
\item \textbf{Bonus (+2 points)}: Exceptional skin texture detail visible throughout video
\item \textbf{Bonus (+1 point)}: Perfect color reproduction and white balance
\item \textbf{Penalty (-3 points)}: Motion blur detected in ANY key facial region (eyes, mouth, teeth) even if brief
\item \textbf{Penalty (-2 points)}: Any visible digital noise or grain
\item \textbf{Penalty (-3 points)}: Unnatural skin smoothing or beauty filter effects
\item \textbf{Penalty (-2 points)}: Inconsistent sharpness between frames (some frames sharp, others blurry)
\end{itemize}

\vspace{0.2cm}
\textbf{EVALUATION PROCESS - FOLLOW THESE STEPS:}

\vspace{0.1cm}
1. First, evaluate each criteria and assign a specific score:
\begin{itemize}[leftmargin=1em, itemsep=0cm, topsep=0.1cm]
   \item Clarity: \_\_\_/35
   \item Stability: \_\_\_/20
   \item Lighting: \_\_\_/20
   \item Artifacts: \_\_\_/15
   \item Occlusion: \_\_\_/10
\end{itemize}

\vspace{0.1cm}
2. Calculate the base score by adding the five scores above:\\
   Base Score = Clarity + Stability + Lighting + Artifacts + Occlusion = \_\_\_

\vspace{0.1cm}
3. Apply bonus/penalty adjustments:
\begin{itemize}[leftmargin=1em, itemsep=0cm, topsep=0.1cm]
   \item List each bonus/penalty with the reason
   \item Calculate adjustment total: \_\_\_
   \item Final Score = Base Score + Adjustment = \_\_\_
\end{itemize}

\vspace{0.1cm}
4. Determine quality tier based on final score

\vspace{0.2cm}
\textbf{MANDATORY OUTPUT FORMAT:}

\begin{verbatim}
STEP 1 - Individual Scores:
Clarity: X/35 (reason for score)
Stability: X/20 (reason for score)
Lighting: X/20 (reason for score)
Artifacts: X/15 (reason for score)
Occlusion: X/10 (reason for score)

STEP 2 - Base Score Calculation:
Base Score = X + X + X + X + X = X/100

STEP 3 - Bonus/Penalty Adjustments:
[List each bonus/penalty with reason and points]
Total Adjustment: +/-X points

STEP 4 - Final Results:
Final Score = X (Base) + X (Adjustment) = X/100
Quality Tier: [Premium/High/Standard/Below Standard]
Critical Issues: [List any issues that prevent premium quality classification]
Motion Blur Check: [Specifically note if eyes/mouth/teeth show any motion blur]
\end{verbatim}

\vspace{0.2cm}
\textbf{IMPORTANT GRADING NOTES:}
\begin{itemize}[leftmargin=0.5em, itemsep=0cm, topsep=0.1cm]
\item Be exceptionally strict with scoring - err on the side of lower scores
\item Only award top scores (90+) for truly exceptional, near-perfect quality
\item Consider that this is for premium training data - standards are higher than typical use
\item Focus on details that would be critical for face restoration model performance
\item Penalize any imperfections that could negatively impact training effectiveness
\item DOUBLE-CHECK your arithmetic at each step to ensure accuracy
\end{itemize}

\vspace{0.1cm}
Please provide a complete evaluation following the exact format above, including all calculation steps.

\end{tcolorbox}

\end{document}